\documentclass[sigconf]{acmart}
\AtBeginDocument{%
  }

\copyrightyear{2026}
\acmYear{2026}
\setcopyright{cc}
\setcctype{by}
\acmConference[MM '26] {Proceedings of the 34th ACM International Conference on Multimedia}{November 10--14, 2026}{Rio de Janeiro, Brazil}
\acmBooktitle{Proceedings of the 34th ACM International Conference on Multimedia (MM '26), November 10--14, 2026, Rio de Janeiro, Brazil}
\acmDOI{10.1145/3767308.3835312}
\acmISBN{979-8-4007-2213-4/2026/11}

\usepackage{bm}
\usepackage[table]{xcolor}
\usepackage{makecell}
\usepackage{balance}
\begin{document}

\title{Did You Steal My Shot? Pioneering Camera Motion Plagiarism Detection in Generative Videos}

\author{Chengguo Zhang}
\email{zhchguo335@163.com}
\affiliation{%
  \institution{Hohai University}
  \city{Nanjing}
  \country{China}
}

\author{Ping Ping}
\correspondingauthor
\email{pingpingnjust@163.com}
\affiliation{%
  \institution{Hohai University}
  \city{Nanjing}
  \country{China}}

\renewcommand{\shortauthors}{Chengguo Zhang \& Ping Ping}

\begin{abstract}
Camera motion often reflects directorial intent and requires professional equipment, making it a high value form of intellectual property. However, generative video models can imitate such high value camera motions with simple prompts, while existing similarity detection methods mainly operate on visual content and fail to capture deeper motion similarity. This is mainly because their training data entangles camera motion with visual content. Moreover, traditional optical flow is insufficient to represent complex camera motions. We therefore build the first benchmark for camera motion analysis, including a motion dataset with \textbf{11} motion styles and evaluation protocols. Furthermore, we propose a motion representation that augments optical flow with vorticity cues from fluid dynamics, thereby better capturing motions. Experiments show that our detector achieves a \textbf{3.02×} improvement in plagiarism detection over the strongest baseline and remains effective on generative videos. We believe our work extends copyright protection beyond static content to dynamic camera motion.
\end{abstract}



\keywords{Camera motion, Generative video, Video representation, Video forensics, Copyright protection}


\maketitle

\section{Introduction}

The rapid proliferation of generative video models, such as Sora 2 \cite{sora2} and Veo 3.1 \cite{veo3}, has significantly lowered the barrier to producing high-quality videos. However, this advent also amplifies concerns regarding copyright protection. Notably, existing discussions on video copyright protection primarily focus on static visual content, such as specific character appearances or scenes, while overlooking a defining attribute of video media: camera motion.

\begin{figure}[t]
\centering
\includegraphics[width=\columnwidth]{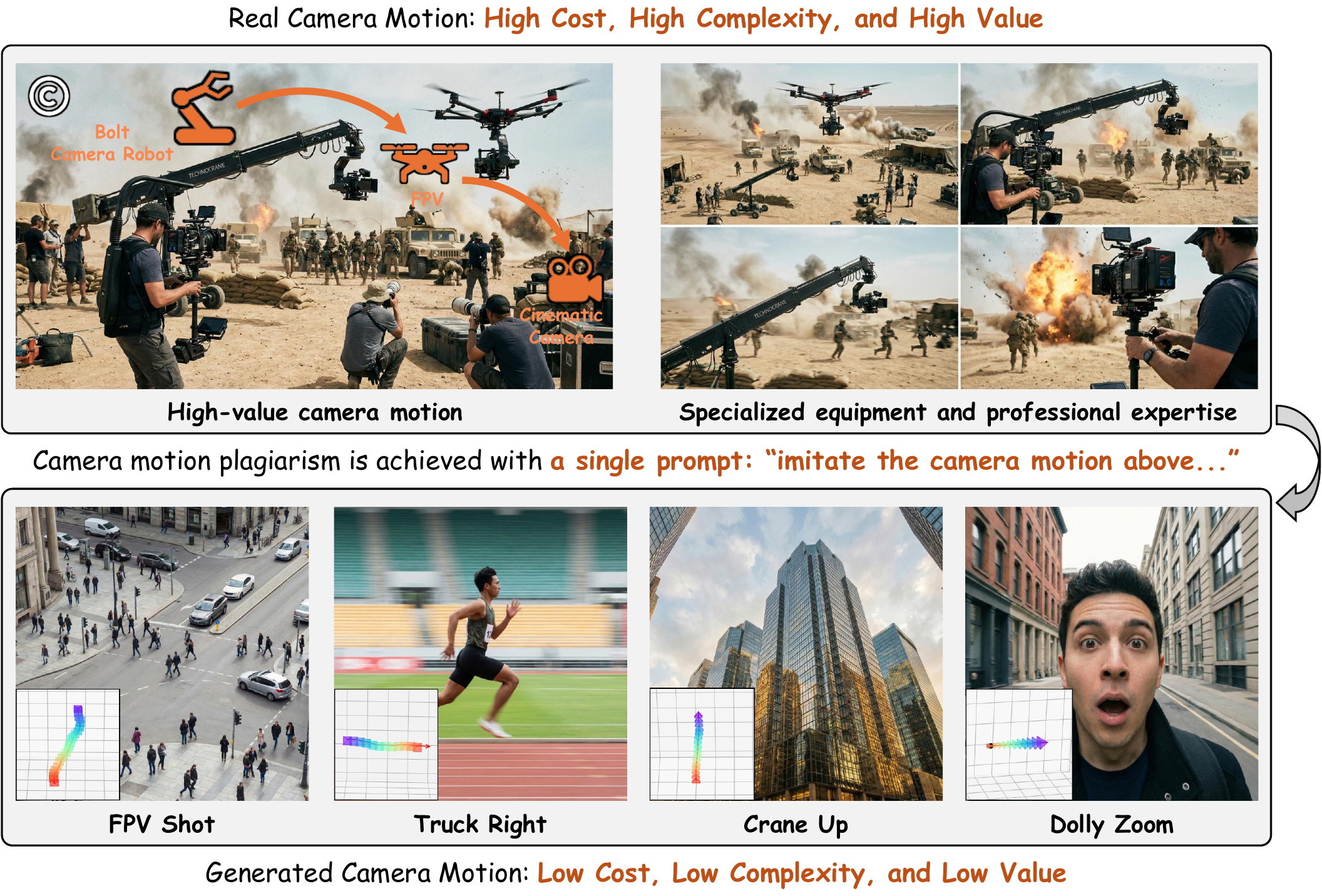}
\caption{Illustration of the Camera Motion Plagiarism. High-value camera motion tends to require specialized equipment and coordination (Top), yet generative videos can easily replicate it with a simple prompt (Bottom), posing a new threat.}
\label{intro}
\end{figure}

A commonly overlooked fact is that camera motion is a high-value creative asset. In professional filmmaking, specific camera motion, such as Dolly Zoom, Bullet Time, or FPV Drone, serves as the director’s narrative signature. Executing these camera motions in the physical world incurs substantial production costs and often requires specialized hardware, such as robotic arms and stabilizers, as well as professional expertise (Fig. \ref{intro}). However, current video generation models make it possible to replicate these complex camera motions at essentially zero cost, which leads to the plagiarism of the creator's intellectual property.

However, detecting such plagiarism poses difficulties that make existing forensic methods ineffective. Traditional techniques such as similarity detection or digital watermarking heavily rely on video content. These methods perform poorly on camera motion plagiarism in generative videos, where visual appearance alters significantly while the temporal pattern of camera motion remains intact. This decoupling between video content and camera motion creates \textit{a forensic blind spot}: the generated video appears visually distinct to evade similarity detection, yet it still mimics the camera motion. This motivates the central research question of our work:

\textbf{\textit{Is it possible to detect camera motion plagiarism?}}

To address this problem, we first analyze why existing forensic methods fail to detect camera motion plagiarism. 1) \textit{Existing video datasets couple motion with content.} Camera motion is an asset independent of video content, yet no existing dataset is designed to isolate camera motion. When models are trained on these video datasets, they tend to exploit shortcut correlations. For example, dolly motion is more common in street scenes, while FPV motion is associated with landscape scenes. As a result, models learn spurious associations between semantics and motion instead of modeling camera motion itself, and fail when generative videos alter the scene. 2) \textit{Ambiguity of optical flow for camera motion.} Optical flow typically encodes motion using Cartesian displacement vectors $(\Delta u, \Delta v)$. However, for complex camera motions, this representation becomes ambiguous as it is highly sensitive to spatial coordinates. Specifically, the same camera motion can yield drastically different optical flow patterns across different image regions, making it difficult for models to capture a unified motion representation. We provide a detailed analysis in Section \ref{section3.1}.

Therefore, we propose CineGuard, the first method designed to detect camera motion plagiarism in generative videos. Given the lack of benchmarks for motion plagiarism detection, we also build CineFlow, a dedicated camera motion dataset. This dataset disentangles camera motion from video content, reducing content bias and improving generalization across scenes. To address the sensitivity of optical flow to spatial coordinates, we introduce vorticity from fluid dynamics as a differential operator. By computing the spatial gradients of the flow field, vorticity extracts higher order geometric features that are independent of absolute pixel locations. It allows the model to capture the representation of camera motion in a spatially invariant manner. As for training strategy, we adopt contrastive learning to group videos with similar camera motion in the feature space. This promotes motion representation learning without requiring explicit labels. Our contributions are as follows:

\begin{itemize}
\item We identify and define camera motion plagiarism for the first time, a critical yet overlooked threat in the generative videos, extending copyright protection from static visual content to dynamic camera
motion.
\item We build CineFlow, the first dedicated dataset and benchmark for camera motion, with 11 motion styles. The dataset isolates camera motion from video content and provides data support for future research on camera motion.
\item We are the first to introduce vorticity into video representation to transform spatially varying optical flow into more consistent features, thereby enhancing the model’s ability to handle complex camera motions.
\item We devise contrastive learning to focus the representations on camera motion rather than video content, enabling zero-shot detection on generative videos.
\end{itemize}

\section{Related Work}

\subsection{Camera Motion}
Camera motion is not merely a physical transformation of viewpoint, but a core element of cinematic language \cite{bordwell2004film,heiderich2012cinematography,hoser2018introduction,camerabook}. Directors often employ various camera motion to express their artistic intent, such as dolly zoom for tension or handheld tracking in chaotic moments.

With the advent of generative video technologies \cite{bu2025bytheway,wang2025av,dong2025talking,yuan2025identity,wang2024recipe}, research in video generation has increasingly moved from content synthesis toward fine-grained camera control. Early diffusion-based models primarily relied on textual prompts (e.g. pan left or zoom in) to implicitly guide camera motion, but natural language is inherently ambiguous and insufficient for describing the precise camera motion. Recent works address this limitation by integrating explicit camera control into diffusion models. MotionCtrl \cite{wang2024motionctrl} and CameraCtrl \cite{he2025cameractrl} incorporate camera pose encoders via adapters or LoRA modules, while DualCamCtrl \cite{zhang2025dualcamctrl} adopts a dual-branch diffusion architecture that leverages depth information to model camera trajectories. CineCtrl \cite{cinectrl} further extends controllability to internal cinematographic parameters such as depth of field, zoom, and shutter speed. In parallel, researchers have begun to formalize camera motion understanding at semantic level, with CameraBench \cite{fang2025camerabench} representing one of the first large-scale benchmarks for evaluating camera motion comprehension in generative video models.

These advances lower the barrier to cinematic creation but also make camera motions easier to imitate. Simple prompts can replicate high-cost camera motions, creating new challenges for copyright protection that remain difficult for existing forensic methods.

\subsection{Generative Video Forensics}
Video forensics has become increasingly important in the generative era, both for detecting synthetic or manipulated content and for supporting copyright protection. Based on the underlying mechanisms, existing approaches can be broadly categorized into proactive defense and passive detection.

Passive detection relies on intrinsic traces left by generative models.
Early work targets GAN artifacts, while recent methods focus on diffusion fingerprints, such as reconstruction based cues and spectral differences \cite{wang2023dire,zheng2025d3,Kim_2025_ICCV}.
For videos, spatiotemporal backbones are used to capture temporal anomalies, including masked autoencoder based features and physical inconsistencies \cite{cai2023marlin,xu2024learning}.

Proactive defense protects video ownership by embedding imperceptible signals during content generation. Recent video watermarking approaches typically inject such signals into the latent space or directly into the spatiotemporal generation pipeline, making them more compatible with diffusion-based video synthesis \cite{jang2024lvmark,huang2025video,fernandez2024video,hu2025videoshield,zhang2024v2a}. Therefore, proactive defense remains focused on protecting video content rather than camera motion.

However, both proactive defense and passive detection primarily target video content rather than dynamic camera motion. When distinctive camera motion is reused while the visual content is substantially altered, these methods often fail to detect the plagiarism.

\section{Methodology}
In this section, we first motivate the construction of a dedicated camera motion dataset and analyze the limitations of Cartesian representations. We then detail the data generation process, explain how vorticity is incorporated, and introduce the use of contrastive learning for similarity detection.

\begin{figure*}[t]
\centering
\includegraphics[width=\textwidth]{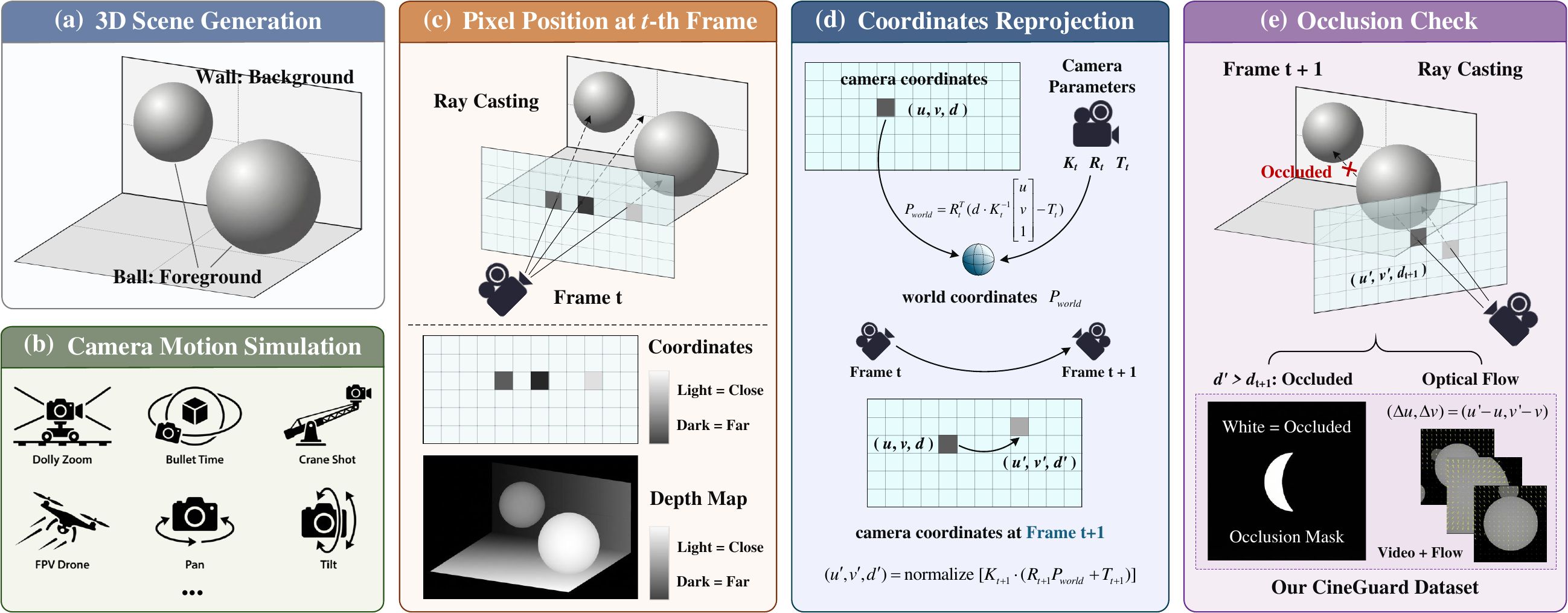}
\caption{Overview of the proposed dataset generation pipeline. (a) and (b): Scenes are constructed using simple geometry and camera motions. (c) 2D coordinates and depth of each pixel at frame $t$ are obtained via ray casting. (d) and (e): Pixel positions at frame $t+1$ are computed via reprojection and occlusion check, and optical flow is derived from coordinate differences.}
\label{dataset}
\end{figure*}

\subsection{Preliminaries and Motivation}

\subsubsection{Limitations of Existing Video Datasets}
Existing video datasets suffer from a fundamental flaw, namely spurious correlations between camera motion and visual content. In natural videos, camera motion is often tightly coupled with specific scenes, such as FPV motion appearing in outdoor landscapes or Dolly in shots occurring in street environments. As a result, forensic models tend to degenerate into content classifiers rather than learning the underlying temporal dynamics of camera motion.

Therefore, a purely geometric dataset is necessary to remove these confounding factors and encourage learning invariant camera motion representations.

\subsubsection{Limitations of Cartesian Optical Flow Representation}
\label{section3.1}
A common way \cite{teed2020raft,huang2024vbench,wang2025shape} to represent motion in videos is through optical flow, which is typically expressed as a dense vector field in Cartesian coordinates:
\begin{equation}
    \mathbf{F}(x, y) = (u, v)
\end{equation}
where $\mathbf{F}(x, y)$ denotes the optical flow vector at pixel $(x, y)$, with $u$ and $v$ representing the horizontal and vertical displacement components. While this representation aligns naturally with translational motion, it exhibits fundamental limitations when modeling complex camera motion such as rotation.

Consider a camera undergoing a rotational motion with a constant angular velocity $\omega$ around a rotation center $\bm{c} = (c_x, c_y)$. The optical flow at pixel coordinate $(x, y)$ can be formulated as:
\begin{equation}
    u(x, y) = -\omega \cdot (y - c_y), \quad v(x, y) = +\omega \cdot (x - c_x)
\end{equation}
The magnitude of the optical flow vector is given by:
\begin{equation}
\begin{aligned}
    |\mathbf{F}(x, y)| &= \sqrt{u(x, y)^2 + v(x, y)^2} \\
    |\mathbf{F}(x, y)| &= \sqrt{[-\omega(y - c_y)]^2 + [\omega(x - c_x)]^2} \\
    |\mathbf{F}(x, y)| &= |\omega| \cdot \sqrt{(x - c_x)^2 + (y - c_y)^2} \\
    |\mathbf{F}(x, y)| &= |\omega| \cdot r
\end{aligned}
\end{equation}
where $r$ is the radial distance from $(x, y)$ to the center. This formulation exposes a key limitation of Cartesian representations: the motion intensity is inherently coupled with spatial coordinates.

Although the angular velocity $\omega$ is a global constant, the magnitude of the resulting optical flow varies with the $r$. As a result, the optical flow field is strongly dependent on pixel location $(x,y)$. Such spatial variability makes it difficult for deep models to learn a compact and consistent representation of the underlying motion dynamics, hindering the detection of camera motion plagiarism across different visual contexts.

Therefore, transforming the representation into a spatially invariant manifold is essential to decouple motion from spatial coordinates and enable robust geometric learning.

\subsection{Generation of CineFlow Dataset}
To reduce spurious correlations between video content and camera motion, we construct CineFlow. As shown in Fig. \ref{dataset}(a), CineFlow models simple 3D scenes composed of basic geometric primitives, including planes and spheres, ensuring that camera motion is the only source of variation. This design prevents reliance on visual appearance and encourages models to focus on motion dynamics.

From a technical perspective, camera motion can be decomposed into six degrees of freedom (DoF), including translational components (truck, pedestal, and dolly) and rotational components (pan, tilt, and roll). In CineFlow, camera trajectories are procedurally synthesized in the 6-DoF space to cover a wide range of camera motion styles, from simple movements to complex motions.

Camera trajectories are generated as smooth temporal sequences. For translation, spline interpolation is applied to ensure continuity of position and velocity over time. For rotation, spherical linear interpolation is used to interpolate rotations on the manifold, preserving constant angular velocity and avoiding artifacts caused by linear interpolation. By combining these 6-DoF components, we simulate 11 distinct camera motion types.

\begin{table}[t]
\centering
\small
\setlength{\tabcolsep}{1.1mm}
\caption{The 11 simulated camera motion types in the CineFlow dataset.}
\label{motion}
\begin{tabular}{l l l}
\toprule
\textbf{Category} & \textbf{Motion Type} & \textbf{DoF} \\
\midrule
Static & Static & None \\
Translation & Truck, Pedestal, Dolly in & $t_x$, $t_y$, $t_z$ \\
Rotation & Pan Tilt, Roll & $r_x+r_y$, $r_z$ \\
\midrule
Composite & \makecell[l]{Bullet Time, Crane, Whip Pan,\\ Dolly Zoom, FPV Drone} & \makecell[l]{$t_x+r_y$, $t_y+t_z$, $r_y+r_z$,\\ $t_z$+zoom, full 6 DoF} \\
\bottomrule
\end{tabular}
\end{table}

\begin{figure*}[t]
\centering
\includegraphics[width=\textwidth]{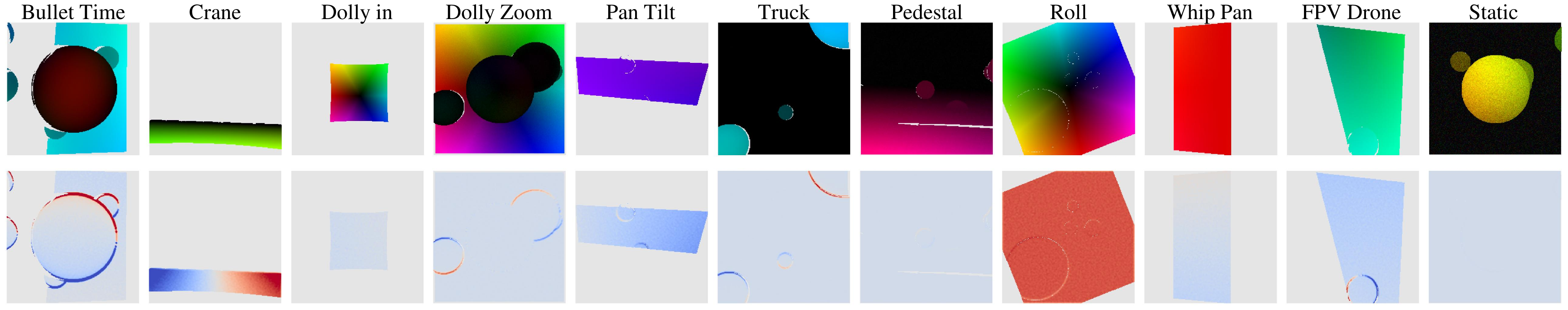}
\caption{Optical flow and vorticity across 11 camera motion styles in CineFlow. First row: optical flow visualizations for one representative clip per motion style, rendered with the standard color coding. Second row: the corresponding vorticity maps computed from the same flow fields.}
\label{dataset-vis}
\end{figure*}

For each camera motion type, we illustrate the motion process using frames $t$ and $t+1$ as an example. At frame $t$, we cast rays from the camera viewpoint into the 3D scene to obtain the position of each point, including its image coordinate $(u, v)$ and distance to the camera (i.e., depth $d$). We refer to this process as ray casting, as shown in Fig. \ref{dataset}(c).

In the previous step, ray casting is used to obtain the position $(u, v, d)$ of a point in frame $t$. However, this position $P^{(t)}_{\mathrm{cam}}$ is defined in the camera coordinates and thus moves with the camera. To compute its position in frame $t+1$, we first transform it into a fixed world coordinates to recover its absolute 3D location $P_{\mathrm{world}}$, and then map it into the camera coordinates of frame $t+1$. As shown in Fig. \ref{dataset}(d), we refer to this process as coordinate reprojection.

Mapping $P^{(t)}_{\mathrm{cam}}$ to $P_{\mathrm{world}}$ requires the $(u, v, d)$ as well as the camera pose at frame $t$, represented by the rotation matrix $R_t$ and translation vector $T_t$. According to the pinhole camera model, the coordinate reprojection is given by:
\begin{equation}
\begin{aligned}
    P^{(t)}_{\mathrm{cam}} = R_t P_{\mathrm{world}} + T_t = d \cdot K^{-1} [u,\, v,\, 1]^{\top}
\end{aligned}
\end{equation}
\begin{equation}
\begin{aligned}
    P_{\mathrm{world}} = R_t^T \left( P^{(t)}_{\mathrm{cam}} - T_t \right)
    = R_t^T (d \cdot K^{-1} [u,\, v,\, 1]^{\top} - T_t)
\end{aligned}
\end{equation}
Given a point in frame $t$, the pixel is first back-projected to 3D using the inverse camera intrinsics $K^{-1}$ and the depth $d$. The resulting point is then transformed into world coordinates by removing the camera translation $T_t$ and applying the inverse rotation $R_t^T$, yielding the absolute 3D position $P_{\mathrm{world}}$.

When the camera moves to frame $t+1$, we transform $P_{\mathrm{world}}$ back into the camera coordinates of frame $t+1$ using the camera parameters $R_{t+1}$ and $T_{t+1}$.
\begin{equation}
    P^{(t+1)}_{\mathrm{cam}} = R_{t+1} P_{\mathrm{world}} + T_{t+1}
\end{equation}
The transformed 3D point is then projected onto the image plane through the intrinsic matrix $K_{t+1}$. According to the pinhole camera model, the corresponding homogeneous image coordinate satisfies:
\begin{equation}
    \lambda
    [u',\, v',\, 1]^{\top}
    =
    K_{t+1} P^{(t+1)}_{\mathrm{cam}}
    =
    K_{t+1}\left(R_{t+1} P_{\mathrm{world}} + T_{t+1}\right),
\end{equation}
where $\lambda$ is the homogeneous scale factor. Applying perspective division yields the image coordinate in frame $t+1$:
\begin{equation}
    [u',\, v',\, 1]^{\top}
    =
    \mathrm{normalize}
    \left[
    K_{t+1} \left( R_{t+1} P_{\mathrm{world}} + T_{t+1} \right)
    \right]
\end{equation}
where $\mathrm{normalize}(\cdot)$ performs perspective division on the homogeneous image coordinate. The depth value at frame $t+1$ is given by the $z$-component of the transformed point:
\begin{equation}
    d' = \left( R_{t+1} P_{\mathrm{world}} + T_{t+1} \right)_z
\end{equation}

Coordinate reprojection provides accurate point-to-point correspondences between consecutive frames. However, reprojection alone does not account for occlusions. As illustrated in Fig. \ref{dataset}(e), a large sphere may occlude a smaller one, in which case the reprojected depth $d'$ is only theoretical. The true depth, denoted as $d_{t+1}$, is obtained via ray casting. We therefore perform occlusion check by comparing $d'$ and $d_{t+1}$. Occluded regions are marked with a mask and excluded during training. A detailed derivation is provided in the Appendix.

Finally, we construct a camera motion dataset of 20k videos, together with optical flow and occlusion masks, covering 11 camera motion types for camera motion plagiarism detection. Additionally, we provide visualizations of CineFlow to illustrate its motion cues. As shown in the first row of Fig. \ref{dataset-vis}, each of the 11 camera motion styles is presented with its optical flow. Optical flow is rendered using the standard color coding, where hue indicates motion direction and saturation reflects motion magnitude. The optical flow is decoupled from video content and reflects only camera motion, making CineFlow particularly suitable for analyzing motion patterns.

\subsection{Vorticity-based Motion Representation}
To address the coordinate dependence of Cartesian optical flow, we introduce a vorticity-based motion representation that captures relative motion variations through spatial differentials of the flow field. Specifically, for a given dense optical flow field $\mathbf{F}(x, y)$, the vorticity $\zeta$ is defined as the curl of the velocity vector:
\begin{equation}
    \zeta = \nabla \times \mathbf{F} \cdot \mathbf{k} = \frac{\partial v}{\partial x} - \frac{\partial u}{\partial y}
\label{vorticity}
\end{equation}
where $\mathbf{F}(x,y) = (u(x,y), v(x,y))$ denotes the optical flow field, $\mathbf{k}$ is the unit vector orthogonal to the image plane.

The core advantage of this representation lies in its spatial invariance. For a pure rotational motion, such as camera roll, the induced optical flow can be written as:
\begin{equation}
    \mathbf{F}(x,y) = \big(-\omega (y - c_y),\; \omega (x - c_x)\big)
\end{equation}
Applying the vorticity operator yields
\begin{equation}
\zeta = \frac{\partial v}{\partial x} - \frac{\partial u}{\partial y} = 2\omega
\end{equation}
This result shows that vorticity is independent of the spatial coordinates $(x, y)$ and the rotation center $(c_x,c_y)$. By contrast, the magnitude and direction of raw optical flow vary with pixel location. Therefore, vorticity provides a spatially invariant representation for camera motion.

\begin{figure}[t]
\centering
\includegraphics[width=0.95\columnwidth]{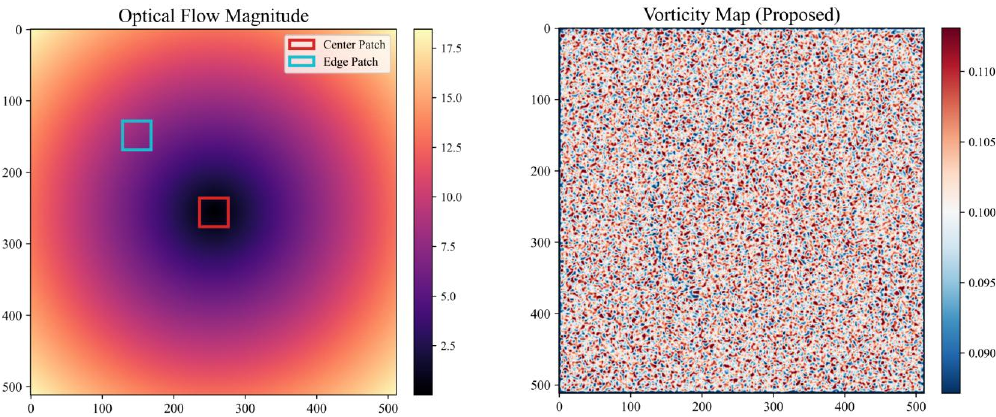}
\caption{Optical flow magnitude (left) and vorticity map (right) under a roll motion.}
\label{vorticity1}
\end{figure}

As illustrated in Fig. \ref{vorticity1}, we select a roll motion and visualize the magnitude of optical flow and the corresponding vorticity map. It is worth noting that the flow magnitude varies strongly with the radial distance from the rotation center, revealing a strong dependence on spatial coordinates. As a result, it fails to provide a unified representation of roll motion, increasing the difficulty of training plagiarism detection models.

\begin{figure}[t]
\centering
\includegraphics[width=0.95\columnwidth]{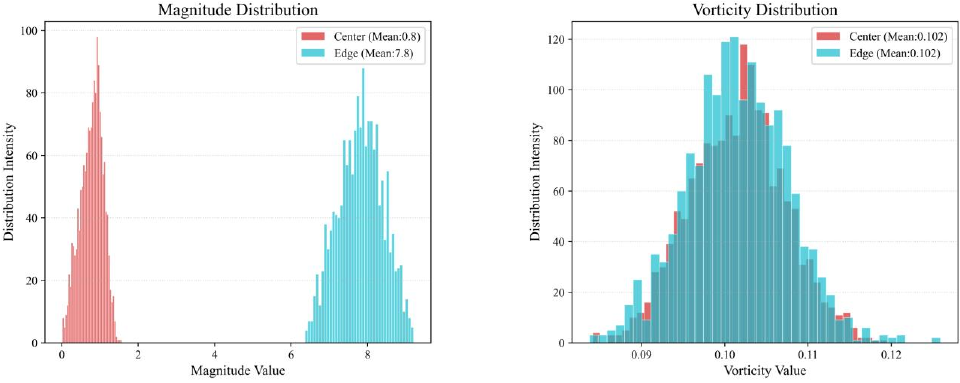}
\caption{Magnitude (left) and vorticity (right) distributions under a roll motion.}
\label{vorticity2}
\end{figure}

In contrast, the vorticity map exhibits a uniform distribution across the image plane, with values remaining relatively consistent regardless of the spatial coordinates. The blue ring near the center arises because vorticity is computed via derivatives, which produce sharp responses at object boundaries, where the flow field exhibits discontinuities. Fig. \ref{vorticity2} further confirms that the vorticity provides a more spatially consistent representation: the center and edge patches exhibit markedly different distributions of flow magnitude, whereas their vorticity distributions nearly overlap.

As shown in the second row of Fig. 3, we further visualize the vorticity of the 11 camera motion styles. Vorticity produces more spatially consistent responses for rotation dominant motions, such as \textit{Roll}, \textit{Bullet Time}, and \textit{Crane}, whereas translation dominant motions show weaker responses. Compared with the optical flow shown in the first row, vorticity suppresses the dependence on spatial location and highlights rotational motion cues more clearly.

To capture complementary motion cues, we construct a joint representation based on optical flow and vorticity. Given a camera motion sequence, we first compute the dense optical flow and apply the vorticity operator to obtain vorticity maps. The flow components and vorticity are then concatenated to form a three-channel motion representation $(u, v, \zeta)$, where $(u, v)$ preserves absolute translational motion, and $\zeta$ captures spatially consistent rotational structures through local flow differentials.

In practice, we observe that vorticity, computed via spatial differentiation, is prone to numerical instability due to noise and motion discontinuities in optical flow. Therefore, to ensure numerical stability, vorticity maps are normalized and clipped to suppress extreme responses during training. For each video clip, the resulting representation sequence is then fed into the detection network.

\begin{figure*}[t]
\centering
\includegraphics[width=0.95\textwidth]{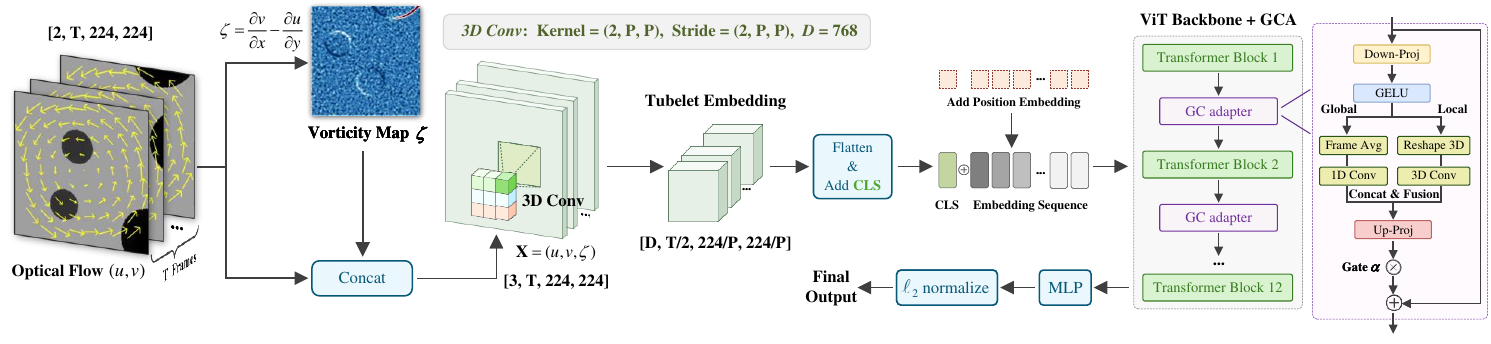}
\caption{Architecture of the motion plagiarism detection network. The input flow $(u,v)$ is first augmented with vorticity $\zeta$. Next, a 3D convolution extracts tubelet embeddings, and a learnable CLS with positional embeddings is added. Then, the embedding sequence is fed into a ViT backbone with GC adapters. Finally, the global embedding is projected.}
\label{network}
\end{figure*}

\vspace{-5pt}
\subsection{Plagiarism Detection Network: CineGuard}
We further propose CineGuard, a transformer-based network for camera motion plagiarism detection, augmented with Global Camera (GC) adapters.

Given a clip of optical flow $\mathbf{F} = (u,v) \in \mathbb{R}^{2 \times T \times H \times W}$, where $T$ denotes the number of frames. As shown in Fig.\ref{network}, the vorticity map $\zeta$ is then computed from the two flow components using Eq. (\ref{vorticity}), followed by normalization and value clipping for numerical stability. We concatenate $(u,v,\zeta)$ along the channel dimension and obtain the network input $\mathbf{X}=(u,v,\zeta)$.

A sequence of tubelet embeddings is then obtained from $\mathbf{X}$ using a 3D convolution \cite{tran2015learning} with temporal tubelet length $2$ and spatial patch size $p$ (set $p=8$). A learnable global embedding is appended, and positional embeddings are added before the sequence is processed by a ViT backbone. Since tubelet embedding produces a different spatiotemporal grid, the pretrained 2D positional embeddings are interpolated in space and replicated along time.

In the ViT backbone, transformer blocks are used to model global spatiotemporal dependencies. Additionally, GC adapters are applied after each transformer block to strengthen camera motion representation. Each adapter has a global branch and a local branch. The global branch averages spatial features per frame and uses a 1D convolution to model camera trajectory cues. The local branch applies a depthwise 3D convolution on the reshaped spatiotemporal grid to capture fine grained motion patterns.

The two branches are fused and injected through a gated residual connection, with the gate initialized to zero. Finally, the global embedding is mapped by an MLP projection head and $\ell_2$ normalized to obtain a compact representation for plagiarism detection.

The network is trained with a contrastive objective. For each flow clip, two augmented views are generated using stochastic spatial cropping, temporal dropout, and horizontal flipping with sign inversion of the horizontal flow component $u$.

Let $(\mathbf{z}, \mathbf{z}^+)$ denote a positive pair obtained from two augmented views of the same clip. $\mathcal{N}$ denote the embeddings from other clips in the same minibatch, which are treated as negatives for $\mathbf{z}$. We use an InfoNCE \cite{oord2018representation} loss:
\begin{equation}
\mathcal{L}=
-\sum_{(\mathbf{z},\mathbf{z}^+)}
\log
\frac{\exp(\mathbf{z}^\top \mathbf{z}^+/\tau)}
{\sum_{\mathbf{z}' \in \{\mathbf{z}^+\}\cup \mathcal{N}}\exp(\mathbf{z}^\top \mathbf{z}'/\tau)}
\end{equation}
where $\tau$ denotes the temperature parameter, and the outer sum is averaged over all positive pairs in the minibatch.

This objective pulls clips with the same motion closer in the embedding space and pushes clips with different motions apart, enabling similarity based motion plagiarism detection.

\section{Experiments}

\subsection{Benchmark Protocol}
For the dataset split, CineFlow is divided into training and test sets with a ratio of 8:2. For the evaluation protocol, we first assess the detector on the CineFlow test split. To evaluate generalization to generative scenarios, we use four open source camera motion models (\textbf{MotionCtrl} \cite{wang2024motionctrl}, \textbf{CameraCtrl} \cite{he2025cameractrl}, \textbf{DualCamCtrl} \cite{zhang2025dualcamctrl} and \textbf{ReCamMaster} \cite{ReCamMaster}) to generate videos from 50 source clips, each with 8 motion styles. The detector is then evaluated by its ability to distinguish these styles. Finally, we test the detector on videos generated by the commercial models: Veo 3.1 \cite{veo3} and Jimeng \cite{jimeng2025} by prompting it to imitate classic camera motions, and test whether such imitation can be detected as plagiarism.

For evaluation metrics, we adopt \textbf{NMI} \cite{NMI} and \textbf{ARI} \cite{ARI} to evaluate the model's ability to distinguish different camera motion styles. In plagiarism detection, candidate videos are ranked by \textbf{cosine similarity} in the embedding space, and \textbf{Precision@5} is reported. For the baselines, since no method is currently available for motion plagiarism detection, we compare against four video foundation models: \textbf{VideoMAE2} \cite{wang2023videomae}, \textbf{ViCLIP} \cite{wang2024internvideo2}, \textbf{DINO v3} \cite{simeoni2025dinov3}, and \textbf{VideoMamba} \cite{VideoMamba}. These models are widely adopted as general purpose backbones and provide strong visual representations.

\subsection{Evaluation on CineFlow Dataset}
We first evaluate whether the proposed detector learns a discriminative representation for camera motion on CineFlow. Specifically, using the joint motion representation as input, the task is to distinguish the 11 camera motion styles. We extract embeddings for $4\times10^3$ test clips and visualize their distribution using t-SNE. A style similarity matrix is further computed by averaging the $\ell_2$ normalized embeddings within each style, and measuring cosine similarity between the resulting prototypes. The matrix is shown as a heatmap to reveal inter style confusions.

\begin{figure}[t]
\centering
\includegraphics[width=\columnwidth]{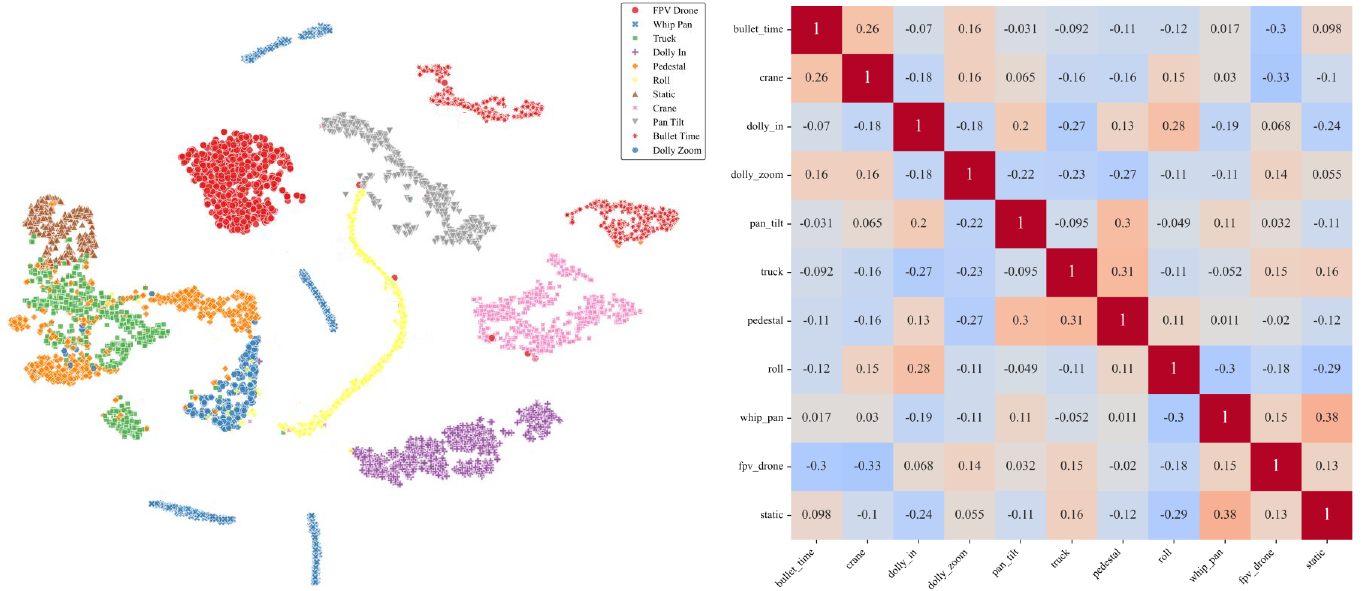}
\caption{Camera motion separability on CineFlow. Left: t-SNE visualization of embeddings extracted from test clips. Right: style similarity heatmap based on cosine similarity.}
\label{tsne+heatmap}
\end{figure}

As shown in Fig. \ref{tsne+heatmap}, the t-SNE visualization forms clear clusters for most motion styles on the test split, reflecting that the learned representations are largely discriminative. However, several translation dominant styles exhibit partial overlap, most notably \textit{static}, \textit{pedestal}, \textit{truck}, and \textit{whip pan}. This is mainly because these styles produce near zero vorticity, and the remaining cue is dominated by flow magnitude, which is insufficient for reliable separation. The similarity heatmap in Fig. \ref{tsne+heatmap} shows a strong diagonal and low off diagonal values, consistent with the t-SNE visualization. Higher similarities are also observed for a few motion styles with similar motion patterns, which explains the residual confusion.

Results on CineFlow dataset suggest that, once camera motion is isolated from visual content, our model learns discriminative representation for distinguishing camera motions. Next, we evaluate whether this ability is generalizable to generative videos.

\begin{figure*}[t]
\centering
\includegraphics[width=0.95\textwidth]{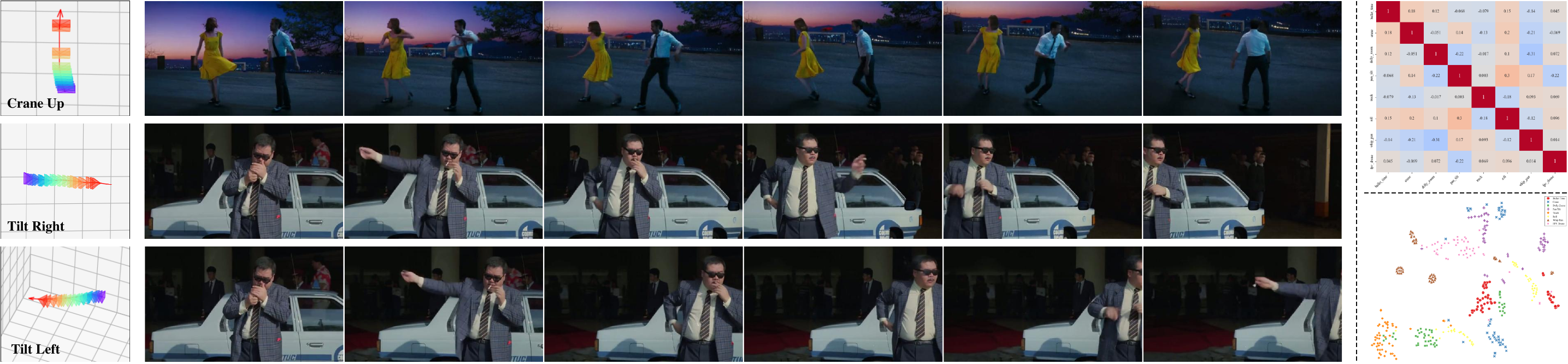}
\caption{Examples and similarity analysis on camera motion models. Left: frame sequences for three motion styles, with the corresponding camera trajectories. Right: the similarity heatmap (top) and the t-SNE visualization (bottom) computed from the generated videos. Additional examples are provided in the Appendix.}
\label{camera}
\end{figure*}

\subsection{Evaluation on Camera Motion Models}
To evaluate generalization, we use videos generated by open source camera motion models. As shown in Fig. \ref{camera}, we select 50 source videos and generate 8 motion styles per video with each model. The detector is evaluated on style separability and compared with the baselines, with results reported in Tab. \ref{nmi} and Tab. \ref{ari}.

Tab. \ref{nmi} reports NMI for clustering the 8 motion styles, where values closer to 1 indicate better agreement with the motion labels. Tab. \ref{ari} reports ARI, where larger values indicate better clustering quality. Although there is still a gap to the ideal scores, our method outperforms all baselines on both metrics across four models. This improvement can be attributed to the fact that our detector learns camera motion dynamics from optical flow and vorticity, rather than relying on visual content cues. In contrast, the baselines remain more sensitive to video content, which makes it difficult for them to cluster videos with the same camera motion reliably.

\begin{table}[t]
\centering
\small
\caption{NMI {\small $\uparrow$} score of clustering on videos generated by camera motion models.}
\label{nmi}
\setlength{\tabcolsep}{1.1mm}
\def\arraystretch{0.9}
\begin{tabular}{ccccc>{\columncolor{blue!10}}c}
    \toprule
    Model & VideoMAE2 & ViCLIP & DINO v3 & VideoMamba & Ours \\
    \midrule
    MotionCtrl & 0.1533 & 0.0715 & 0.1261 & 0.2024 & 0.6470 \\
    CameraCtrl & 0.1692 & 0.0598 & 0.0951 & 0.1748 & 0.5755 \\
    DualCamCtrl & 0.1408 & 0.0722 & 0.1137 & 0.1515 & 0.5913 \\
    ReCamMaster & 0.1524 & 0.0783 & 0.1496 & 0.1773 & 0.6201 \\
    \bottomrule
\end{tabular}
\end{table}

\begin{table}[t]
\centering
\small
\caption{ARI {\small $\uparrow$} score of clustering on videos generated by camera motion models.}
\label{ari}
% \normalsize
\setlength{\tabcolsep}{1.1mm}
\def\arraystretch{0.9}
\begin{tabular}{ccccc>{\columncolor{blue!10}}c}
    \toprule
    Model & VideoMAE2 & ViCLIP & DINO v3 & VideoMamba & Ours \\
    \midrule
    MotionCtrl & -0.0762 & -0.1094 & -0.0868 & -0.0316 & 0.3739 \\
    CameraCtrl & -0.0847 & -0.1251 & -0.1033 & -0.0539 & 0.3486 \\
    DualCamCtrl & -0.0735 & -0.1327 & -0.0979 & -0.0101 & 0.3844 \\
    ReCamMaster & -0.0880 & -0.1288 & -0.0942 & -0.0672 & 0.3717 \\
    \bottomrule
\end{tabular}
\end{table}

The remaining gap to the ideal scores is expected. Generator artifacts and scene specific motion patterns introduce additional variations within the same style, making clusters less compact and increasing confusion between motion styles. Despite these challenges, our detector still achieves strong clustering performance.

\subsection{Evaluation on Commercial Models}
\begin{figure}[t]
\centering
\includegraphics[width=\columnwidth]{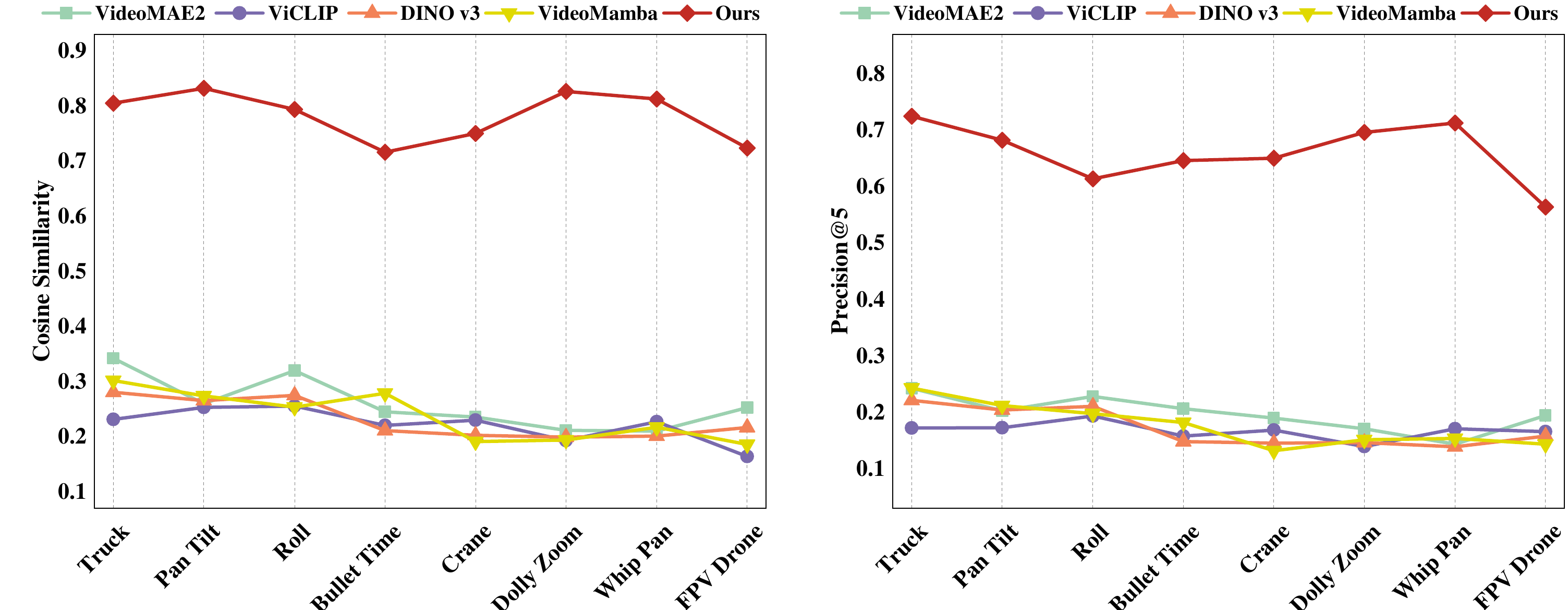}
\caption{Performance of plagiarism detection on generative videos. Left: cosine similarity between videos. Right: Precision@5 for retrieving videos with the same motion label.}
\label{veo}
\end{figure}

\begin{figure*}[t]
\centering
\includegraphics[width=0.95\textwidth]{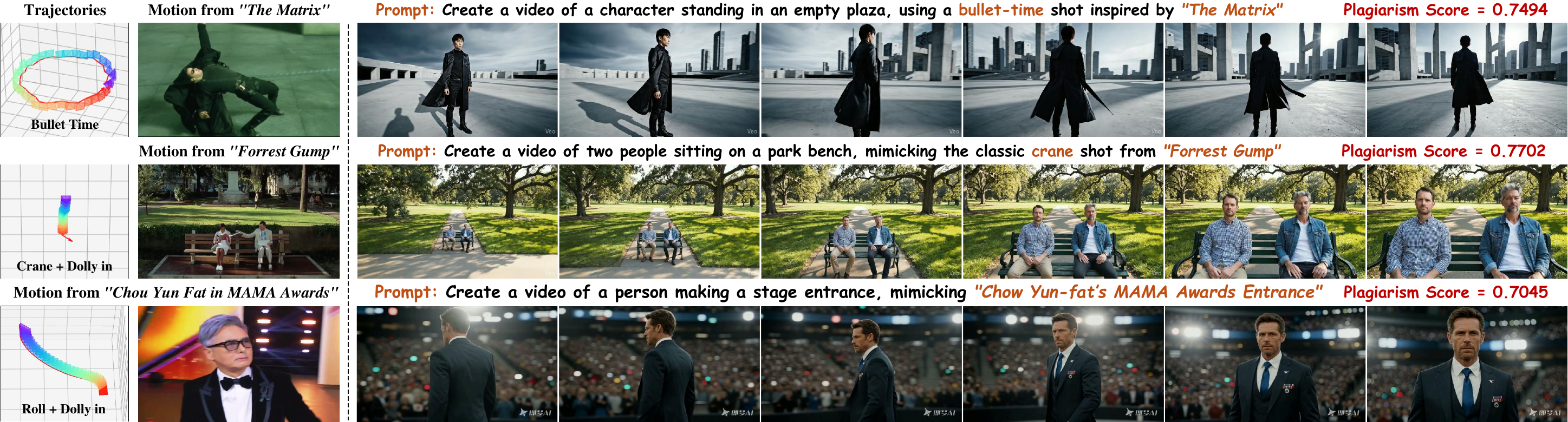}
\caption{Motion plagiarism detection on commercial generative videos. Given a reference camera motion (left), Veo 3.1 and Jimeng are prompted to imitate the motion while generating a new scene (middle). CineGuard outputs a motion plagiarism score based on cosine similarity (right). Additional examples are provided in the Appendix.}
\label{camera2}
\end{figure*}

To assess practical applicability, we further evaluate the detector on commercial generative video tools, Veo 3.1 and Jimeng. Specifically, we select eight classic camera motions, which are widely used in filmmaking. As shown in Fig. \ref{camera2}, Veo is prompted to imitate a given reference motion while generating a new video. For each motion, 20 imitation videos are produced. Camera motion plagiarism is then evaluated using cosine similarity and Precision@K (set $K=5$), and compared with the baselines, with results shown in Fig. \ref{veo}.

As shown in the left plot of Fig. \ref{veo}, our detector yields higher cosine similarity scores, indicating it captures motion similarity despite variations in video content. However, FPV Drone is more challenging to detect since it combines multiple motion styles and is more easily confused with other styles. The Precision@K results in the right plot of Fig. \ref{veo} show consistent gains over the baselines, suggesting fewer false positives in the top ranked results. These two metrics suggest our CineGuard is promising for camera motion plagiarism detection on commercial generative videos.

\begin{figure*}[t]
\centering
\includegraphics[width=\textwidth]{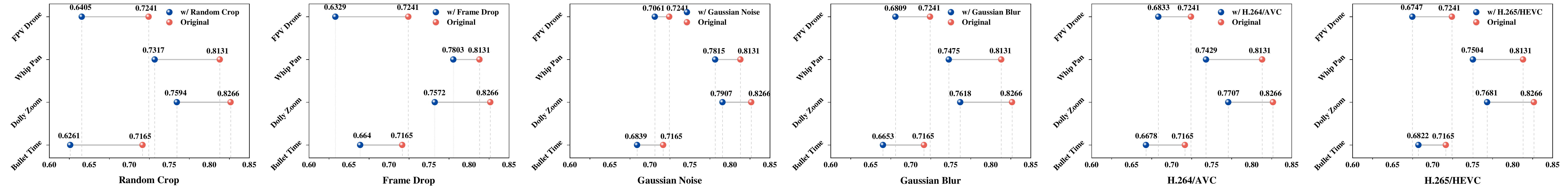}
\caption{Robustness under common post processing attacks. The figure includes Random Crop, Frame Drop, Gaussian Noise, Gaussian Blur, H.264/AVC, and H.265/HEVC, for four representative camera motion styles.}
\label{attack}
\end{figure*}

\subsection{Robustness of CineGuard}
To evaluate robustness under practical attacks, we further conduct experiments with six common post processing attacks: Random Crop, Frame Drop, Gaussian Noise, Gaussian Blur, H.264/AVC compression, and H.265/HEVC compression. These attacks cover temporal disturbance, spatial distortion, additive noise, blur, and lossy video compression.

As shown in Fig. \ref{attack}, we select generated videos with four representative camera motion styles and visualize the change in cosine similarity before and after the attacks. It can be observed that the geometric attacks, especially Random Crop and Frame Drop, cause the largest degradation, while noise based attacks have a relatively smaller effect. Although the detector shows some performance drop under these attacks, it still maintains a relatively high detection rate. This robustness mainly benefits from the augmentation strategy used during training, which improves the model's tolerance to spatial and temporal attacks.

\begin{table}[t]
\centering
\small
\caption{Cosine similarity {\small $\uparrow$} under ablation on CineFlow.}
\label{abl-cineflow}
\setlength{\tabcolsep}{1.1mm}
\def\arraystretch{0.9}
\begin{tabular}{cccccc}
    \toprule
    Dataset & VideoMAE2 & ViCLIP & DINO v3 & VideoMamba & Ours \\
    \midrule
    w/o CineFlow  & 0.2594 & 0.2228 & 0.2367 & 0.2311 & 0.3621 \\
    \rowcolor{blue!10}
    w/ CineFlow & 0.5427 & 0.4783 & 0.5032 & 0.5265 & 0.7828 \\
    \bottomrule
\end{tabular}
\end{table}

\begin{table}[t]
\centering
\small
\caption{Cosine similarity {\small $\uparrow$} under ablation on vorticity.}
\label{abl-vorticity}
\setlength{\tabcolsep}{1.1mm}
\def\arraystretch{0.9}
\begin{tabular}{cccccc}
    \toprule
    Dataset & VideoMAE2 & ViCLIP & DINO v3 & VideoMamba & Ours \\
    \midrule
    w/o vorticity  & 0.5427 & 0.4783 & 0.5032 & 0.5265 & 0.5371 \\
    \rowcolor{blue!10}
    w/ vorticity & 0.6383 & 0.5541 & 0.5806 & 0.6074 & 0.7828 \\
    \bottomrule
\end{tabular}
\end{table}

\subsection{Ablation Study}
We ablate the CineFlow dataset, the vorticity channel to assess their respective contributions.

\subsubsection{Ablating the CineFlow Dataset}
To evaluate the contribution of CineFlow to camera motion understanding, we compare training on CineFlow with training on a general purpose video dataset, Kinetics-400 \cite{kay2017kinetics}, and compare it with the version trained on CineFlow. For each baseline method, we compare the original pretrained checkpoint with the version retrained on CineFlow. Results are reported in Tab. \ref{abl-cineflow}.

Results show that all baselines consistently improve after being retrained on CineFlow, especially VideoMAE2. Such ablation study suggests CineFlow is explicitly designed to isolate motion from video content, providing cleaner motion cues for representation learning. Moreover, the consistent gains across all baselines demonstrate that CineFlow is not only beneficial to our method, but also serves as a generally effective dataset for improving camera motion understanding.

\subsubsection{Ablating the Vorticity Channel}
We evaluate the effect of vorticity with two complementary settings. First, we remove $\zeta$ and train our detector using only optical flow $(u,v)$. Second, based on the previous ablation setting, we further equip each baseline with the vorticity channel $\zeta$ and retrain it on CineFlow under the same protocol. Results in Tab. \ref{abl-vorticity} show that removing $\zeta$ reduces detection performance, while adding $\zeta$ improves the original performance of the baselines. It can be explained by the fact that raw optical flow mainly captures local displacement magnitude and direction, while vorticity explicitly characterizes local rotational structure in the motion field. This confirms that the vorticity provides complementary motion cues beyond raw optical flow.

\section{Conclusion}
In this work, we study camera motion plagiarism in generative videos, an important yet overlooked problem. We further analyze why existing video similarity methods struggle with this task. Motivated by these findings, we introduce CineFlow, a dedicated benchmark for camera motion analysis. We further propose CineGuard, which incorporates vorticity to strengthen motion representations for plagiarism detection. We believe our method provides a promising solution for camera motion plagiarism detection, and extends copyright protection from static content to dynamic motion.

\begin{acks}
  This work is supported by the National Natural Science Foundation of China under Grant 62402182 and 61902110.
\end{acks}

\bibliographystyle{ACM-Reference-Format}
\balance
\bibliography{sample-base}
\end{document}